\documentclass[journal]{IEEEtran}

\usepackage{times}
\usepackage{epsfig}
\usepackage{graphicx}
\usepackage{amsmath}
\usepackage{amssymb}
\usepackage{multirow}
\usepackage{cite}
\usepackage{comment}

\ifCLASSINFOpdf

\else

\fi

\begin{document}

\title{RZSR: Reference-based Zero-Shot Super-Resolution with Depth Guided Self-Exemplars}

\author{Jun-Sang Yoo,
        Dong-Wook Kim,
        Yucheng Lu,
        and~Seung-Won Jung, \emph{Senior Member, IEEE}
\thanks{\textit{Corresponding author: Seung-Won Jung.}}
\thanks{J.-S. Yoo, D.-W. Kim, and S.-W. Jung are with the Department
of Electrical and Electronic Engineering, Korea University, Seoul,
Korea. Y. Lu is with the Education and Research Center for Socialware IT, Korea University, Seoul, Korea (e-mail: junsang7777@naver.com; spnova12@gmail.com; yucheng.l@outlook.com; swjung83@korea.ac.kr)}
}

\maketitle

\begin{abstract}
Recent methods for single image super-resolution (SISR) have demonstrated outstanding performance in generating high-resolution (HR) images from low-resolution (LR) images. However, most of these methods show their superiority using synthetically generated LR images, and their generalizability to real-world images is often not satisfactory. In this paper, we pay attention to two well-known strategies developed for robust super-resolution (SR), i.e., reference-based SR (RefSR) and zero-shot SR (ZSSR), and propose an integrated solution, called reference-based zero-shot SR (RZSR). Following the principle of ZSSR, we train an image-specific SR network at test time using training samples extracted only from the input image itself. To advance ZSSR, we obtain reference image patches with rich textures and high-frequency details which are also extracted only from the input image using cross-scale matching. To this end, we construct an internal reference dataset and retrieve reference image patches from the dataset using depth information. Using LR patches and their corresponding HR reference patches, we train a RefSR network that is embodied with a non-local attention module. Experimental results demonstrate the superiority of the proposed RZSR compared to the previous ZSSR methods and robustness to unseen images compared to other fully supervised SISR methods.
\end{abstract}

\begin{IEEEkeywords}
Deep learning, image super-resolution, reference-based super-resolution, zero-shot super-resolution
\end{IEEEkeywords}

\IEEEpeerreviewmaketitle

\section{Introduction}
\IEEEPARstart {I}{mage} super-resolution (SR) focuses on recovering a high-resolution (HR) image from its corresponding low-resolution (LR) image. Although significant endeavors have been made especially for single image SR (SISR), its inherent ill-posed characteristics make SISR still challenging. As with other computer vision tasks, convolutional neural networks (CNNs) are currently dominating SISR. Despite the outstanding progress of recent methods~\cite{dong2014learning,kim2016accurate,lim2017enhanced,ledig2017photo,he2020tmm,zhang2021tmm}, the SR performance is often unsatisfactory when the training and test images have different characteristics. In other words, most SISR methods are not yet robust to in-the-wild images.

\begin{figure}[t]
    \centering
    \includegraphics[width=\linewidth]{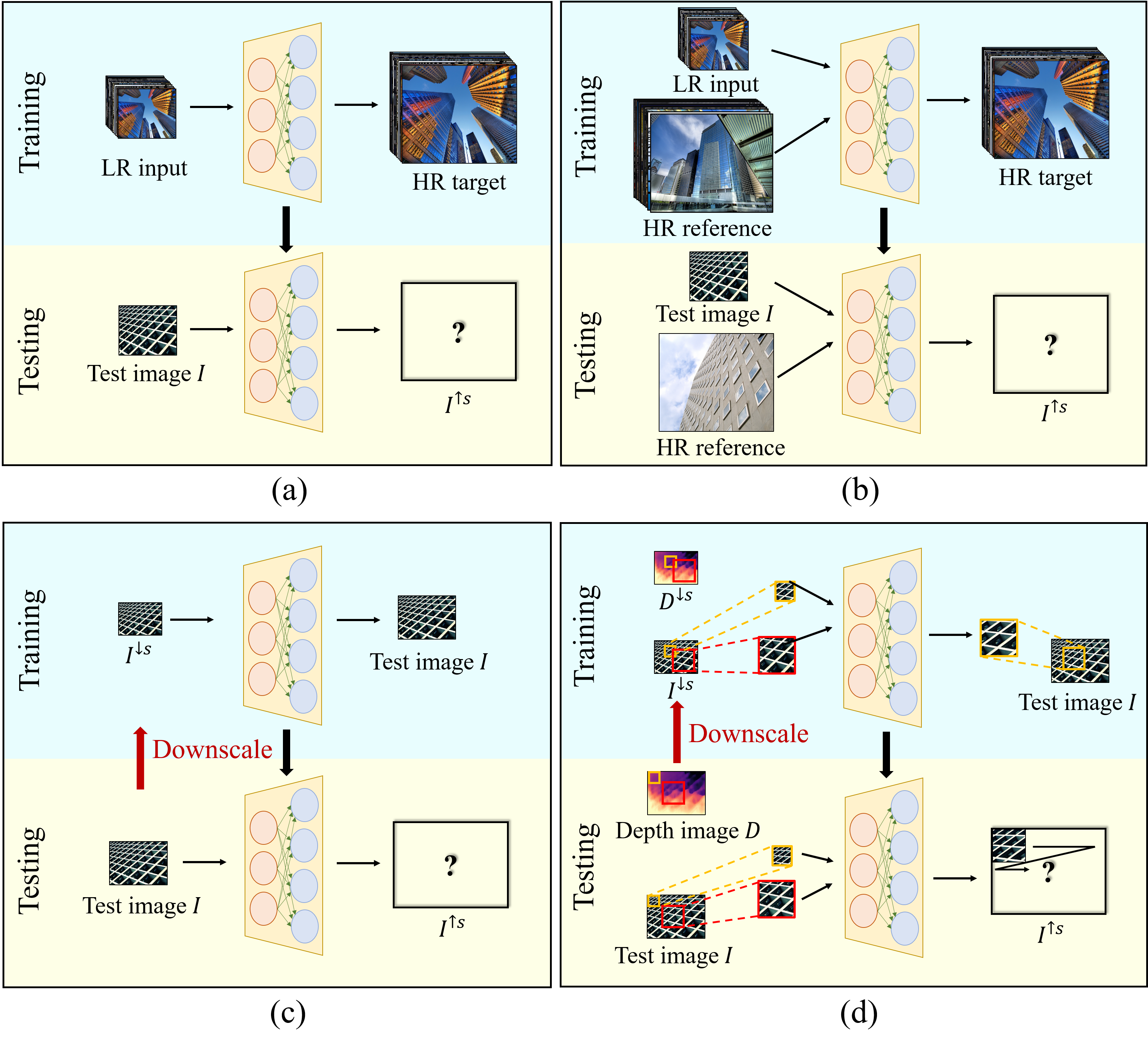}
    \caption{Comparison of SR methods: (a) Conventional fully supervised SR, (b) RefSR, (c) ZSSR, and (d) RZSR. }
    \label{fig:intro}
\end{figure}

Meanwhile, reference-based SR (RefSR) methods have been proposed to improve the robustness of SISR~\cite{zheng2018crossnet,zhang2019image,yang2020learning}. Compared to classic SISR, RefSR recovers rich textures and high-frequency details by using high quality and/or HR reference images as shown in Figs.~\ref{fig:intro}(a) and (b). Many RefSR methods have addressed how to obtain reliable reference images ~\cite{cui2020towards,mei2020image,huang2015single} or use less similar reference images robustly ~\cite{shim2020robust,zhang2019image,xie2020feature}. However, it is not guaranteed that the reference images that can be obtained at the test stage will have a similar level of quality to the reference images used in the training stage. Therefore, RefSR alone cannot be the best solution for real-world image SR.   

To overcome the limitations of the aforementioned SR methods that rely on external supervision, Shocher \textit{et al.}~\cite{shocher2018zero} proposed zero-shot SR (ZSSR), which trains an SR network using only a single LR input image. Specifically, an image-specific SR network is trained using the input LR image and its further downsampled version as illustrated in Fig.~\ref{fig:intro}(c). Consequently, the internal recurrence of information inside a single image can be implicitly learned, and thus SR can be robustly performed using a simpler network architecture. However, this simple network has insufficient modeling capacity due to its limited receptive field as well as its lack of explicit handling of image self-similarity.

\begin{figure*}[t]
\begin{center}
\includegraphics[width=0.99\linewidth]{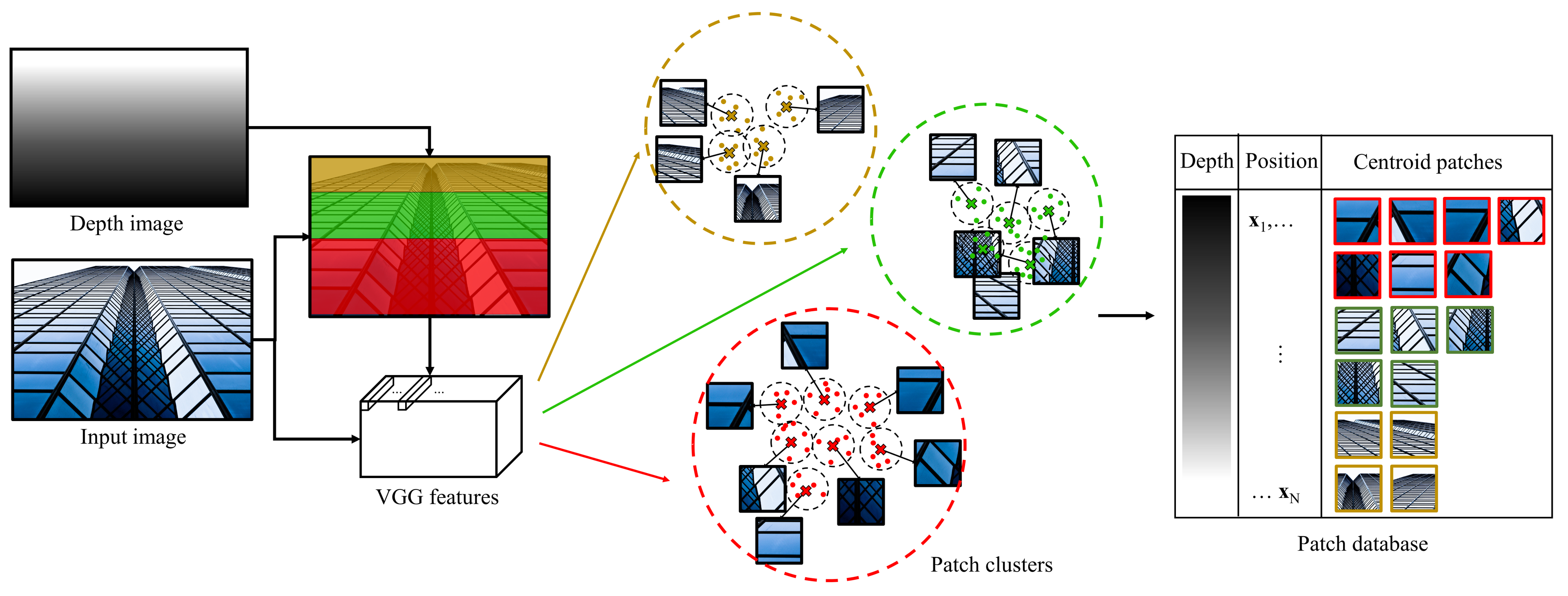}
\end{center}
   \caption{Illustration of the internal patch database construction process: From the depth map estimated from the input image, the input image is first divided into multiple segments according to depth ranges. K-medoids clustering is then performed for each segment using the VGG features extracted from the input image. Patch database contains centroid patches after K-medoids clustering and their corresponding center positions and depth values.}\label{fig:patchgen}
\end{figure*}

In this paper, to take advantage of both RefSR and ZSSR, we present a new SR method called reference-based zero-shot super-resolution (RZSR), as shown in Fig.~\ref{fig:intro}(d). RZSR not only uses image patches from the same input image as ZSSR but also uses HR reference patches like RefSR. However, unlike RefSR, which uses HR patches from external images, our RZSR also derives HR reference patches from the same input image. In this manner, we can exploit the internal patch recurrences more explicitly and inherit high-frequency details from HR reference patches. Specifically, we first cluster image patches according to their predicted depth values and obtain representative patch samples. For each LR patch, we then find its closest representative patch located closer than itself. In this way, rich-textured reference patches across scales can be found efficiently for training and testing. Moreover, a dedicated RefSR network is designed to fully exploit reference patches using a non-local attention module.

Our contributions are summarized as follows:
\begin{itemize}
    \item We present a simple but effective method for constructing an internal HR patch database for RefSR of the input image. Moreover, we propose an efficient method of assigning a relevant HR reference patch for each LR patch using depth information.
    \item We design a network architecture for RZSR that can take full advantage of HR reference patches, thus leading to significant performance improvement compared to its reference-free version. 
    \item RZSR shows better robustness to unseen images than fully-supervised SR methods and outperforms other ZSSR-based methods, particularly on real-world challenging LR images.
    
\end{itemize}

\section{Related Work}

Because SISR has been extensively studied in the last decades, an interested reader can refer to recent articles~\cite{wang2020review, Anwar2020review} for a historical and categorical literature review. Here, we review only the related studies on RefSR, ZSSR, and image self-similarity, which are closely related to our RZSR.

\subsection{Reference-based Super-Resolution} 
Unlike classic SISR, RefSR relies on additional patches or images with HR textures and details, defined simply as a reference image, that can help SR of the given LR image. Consequently, obtaining a high-quality reference image is of utmost importance for RefSR. In many prior studies, the reference image is obtained externally. For example, the reference image is selected from temporally adjacent video frames~\cite{caballero2017real,liu2011bayesian}, retrieved from web images~\cite{yue2013landmark}, or obtained from different viewpoints~\cite{zheng2018crossnet}. Because the reference images that can be obtained by these methods cannot be precisely aligned with the given LR image, special attention needs to be given to enhance its usefulness. 

To this end, Zheng \textit{et al.}~\cite{zheng2018crossnet} proposed CrossNet, which performs feature alignment using multi-scale optical flow vectors. 
Zhang \textit{et al.}~\cite{zhang2019image} aligned the LR and reference images by extracting their VGG features~\cite{simonyan2014very} and applying local patch matching. Yang \textit{et al.}~\cite{yang2020learning} proposed a texture transfer network that searches for the relevant textures from the reference image and transfers them to LR features to learn a more powerful feature representation. Xie \textit{et al.}~\cite{xie2020feature} introduced a feature match and swap module such that texture features of the reference image can be directly swapped with the features from the LR input image. Toward content-independent RefSR, Yan \textit{et al.}~\cite{cui2020towards} built a universal reference pool consisting of a large number of key features. 

\subsection{Zero-Shot Super-Resolution} 
Zero-shot learning (ZSL) aims to predict labels unseen during training. Specifically, ZSL does not rely on pre-training or any external image samples but uses only the input image for training and testing. Since ZSL is free from the domain gap between training and test datasets, it has been extensively studied for real-world image restoration problems ~\cite{shocher2018zero, cheng2020zero, bell2019blind}. 
In a pioneering work, Shocher \textit{et al.}~\cite{shocher2018zero} introduced a ZSL-based SR method called ZSSR, which does not rely on pre-training or any external image samples for SR of the given image. Specifically, they trained an image-specific CNN by using HR-LR patch pairs obtained only from the given image and its downsampled version. ZSSR showed superiority over fully-supervised SR methods on real-world images for which the image acquisition process is unknown or non-ideal. It can also handle SR to any size and any aspect ratio.

To advance ZSSR, Bell \textit{et al.}~\cite{bell2019blind} proposed a generative adversarial network (GAN)~\cite{goodfellow2014generative}-inspired ZSSR method named KernelGAN. Specifically, they predicted an image-specific downscaling kernel using a generator such that a discriminator cannot distinguish between the patch distribution of the original and downscaled images. However, since KernelGAN is decoupled with the SR network, it can lead to suboptimal SR performance. To solve this problem, Cheng \textit{et al.}~\cite{cheng2020zero} coupled the degradation simulation network and the super-resolution network. To train these coupled networks, they generated an internal unpaired HR-LR patch dataset using depth information and designed a bi-cycle training strategy. Emad \textit{et al.}~\cite{emad2021} presented a similar method, called DualSR, using GAN and cycle consistency loss. Rather than using depth information, they used patches of different sizes to train the degradation simulation network and the super-resolution network.

\subsection{Self-similarity in SISR} 

Many prior studies have demonstrated that in-scale and cross-scale patch similarity exists in natural images~\cite{wan2018self,dai2019second,shim2020robust,mei2020image}. This similarity, called self-similarity, has been widely used to enable internal SR that builds a training dataset from the input image itself. The pioneering work of Glasner \textit{et al.}~\cite{glasner2009super} exploited self-similarity through in-scale and cross-scale patch matching and integrated example-based SR constraints with conventional SR constraints. Another representative work by Huang \textit{et al.}~\cite{huang2015single} found more expressive self-exemplars by expanding the internal search space using geometric transformations. Chen \textit{et al.}~\cite{chen2017self} characterized self-similarity prior in the transform domain using the local structure-adaptive transform, while Chantas \textit{et al.}~\cite{chantas2020self} introduced a variational approach with an observation that the self-similarity follows a heavy-tailed distribution.

Most recent learning-based SR networks have exploited self-similarity using a non-local neural network~\cite{wang2018non}. For example, Dai \textit{et al.}~\cite{dai2019second} presented a deep residual network that incorporates non-local operation to capture long-distance spatial contextual information. Shim \textit{et al.}~\cite{shim2020robust} introduced a self-similarity SR framework that estimates dynamic offsets for deformable convolution. Mei \textit{et al.}~\cite{mei2020image} proposed a cross-scale non-local attention module that explores cross-scale feature similarities. However, all of these methods rely on external datasets and can only exploit self-similarity within a limited receptive field.

We exploit self-exemplars as others~\cite{glasner2009super,huang2015single} and non-local networks as others~\cite{dai2019second,shim2020robust,mei2020image} for SR, but in a novel direction. By using the self-exemplars found from the same image as reference patches and feeding them as additional input to our SR network embodied with a non-local attention module, we can fully exploit the self-exemplars under the ZSL framework. In short, we propose RZSR, which is the integration of RefSR, ZSSR, and self-similarity. In order to fully exploit the self-similarity for ZSSR, we first construct an internal HR patch database and apply cross-scale patch matching using depth information. The HR patches retrieved from the internal database are then used as reference for training our image-specific RefSR network. The details of RZSR will be provided in the next section.

\begin{figure*}[t]
\begin{center}
\includegraphics[width=\linewidth,page=1]{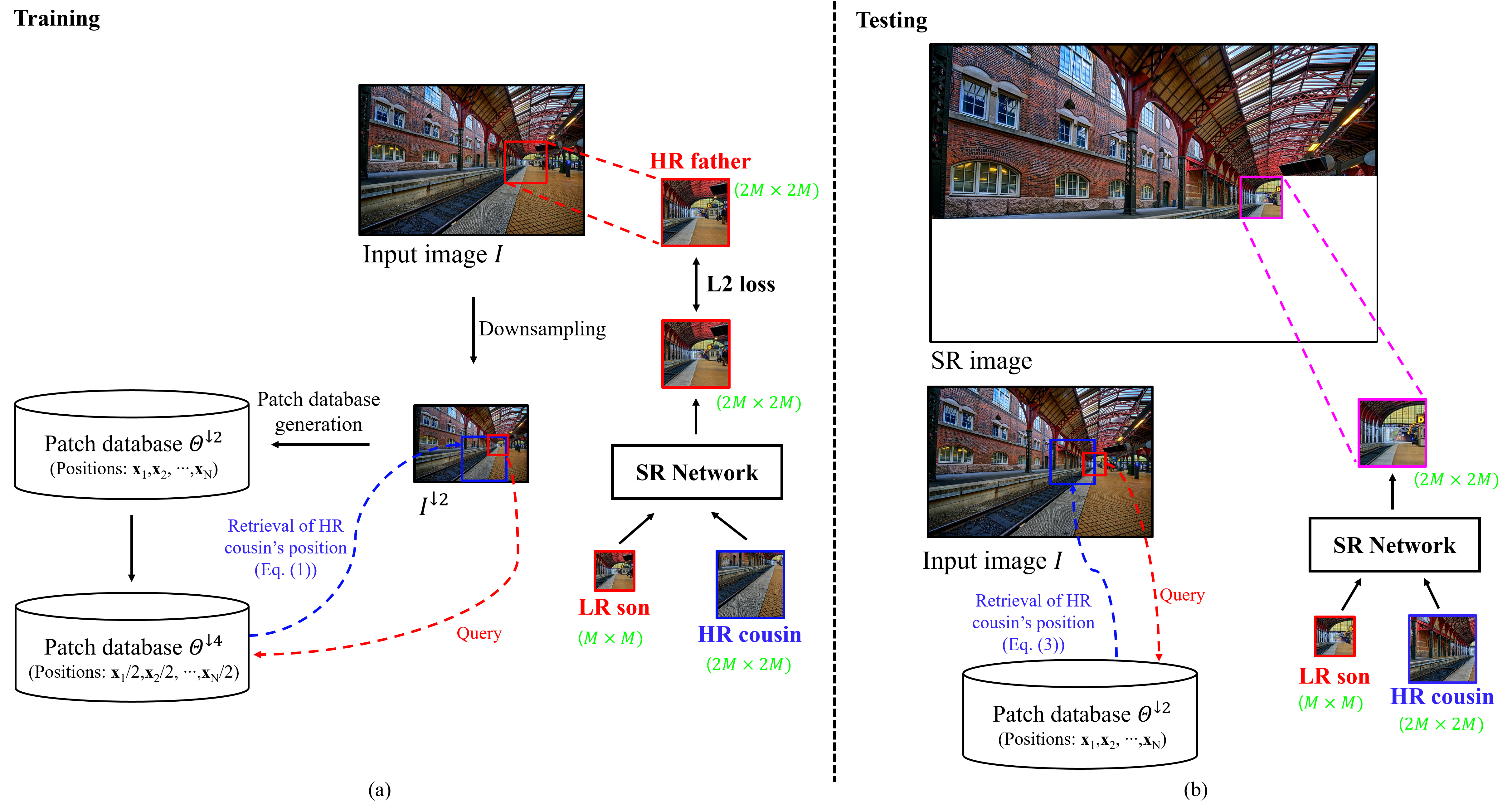}
\end{center}
   \caption{Illustration of (a) the training stage and (b) testing stage. In the training stage, input image $I$ is first downsampled, and the patch database $\Theta^{ \downarrow 2}$ is then constructed from the downsampled image ${I^{ \downarrow 2}}$, as explained in Sec.~\ref{sec:patchgen}. Each image patch in $\Theta^{ \downarrow 2}$ is treated as LR son, and its corresponding HR patch in $I$ is treated as HR farther. LR son is queried to $\Theta^{ \downarrow 4}$, and its HR cousin's position is retrieved, as explained in Sec.~\ref{sec:patchretrieve}. Given LR son and HR cousin, SR network is trained using HR father as a target signal, as explained in Sec.~\ref{sec:network}. In the testing stage, each image patch in $I$ is treated as LR son, and its HR cousin's position is retrieved from $\Theta^{ \downarrow 2}$. SR output image is reconstructed by aggregating patch-wise SR results, as explained in Sec.~\ref{sec:inference}. The size of each patch is marked in green. }\label{fig:patchmatch}
\end{figure*}

\section{Proposed method}

\subsection{Construction of Internal Patch Database} \label{sec:patchgen}
Similar to other RefSR methods, RZSR requires a high-quality reference image for high-performance SR. Motivated by the previous internal SR methods~\cite{glasner2009super, huang2015single}, we seek high-quality reference patches from the input image through cross-scale patch matching. However, exhaustive patch matching can introduce unacceptable computational costs, especially because RZSR needs to perform network training at the test stage using a given LR image. Therefore, we develop a simple but effective patch database construction method that uses depth information.

Fig.~\ref{fig:patchgen} illustrates the construction process of the internal patch database. Here, we assume that the depth map, which is aligned with the LR image, is given. This depth map can be obtained using monocular depth estimation techniques~\cite{godard2019digging, bhat2021depth} or depth sensors with an additional image alignment process~\cite{janoch2013category,silberman2012indoor,xiao2013sun3d}. We construct the internal reference patch database based on two intuitions: First, for a given LR patch, its corresponding rich textures and details are highly likely to be located in regions closer than itself. Second, redundant patches exist in natural images, so a few representative patches are sufficient for use as matching candidates. 

Based on these two intuitions, we first divide the LR image according to depth ranges, where we uniformly divide the whole depth range into $D$ ranges. We then collect image patches according to the depth value at the center and finally obtain representative image patches for each depth range by applying $k$-medoids clustering~\cite{park2009simple} to the VGG features~\cite{simonyan2014very,johnson2016perceptual} of the image patches. The experimental results showed that using the center pixel value was more advantageous for the subsequent clustering process than using the average or median value. The resultant patch database thus consists of the representative patches and their center positions and depth values, as depicted in Fig.~\ref{fig:patchgen}.

\subsection{Reference Patch Retrieval} \label{sec:patchretrieve}
Similar to the previous ZSSR methods~\cite{shocher2018zero, bell2019blind}, we can obtain LR-HR patch pairs from the original LR image and its downsampled version. We call the corresponding LR-HR patches \textit{LR son} and \textit{HR father}, respectively. Toward RZSR, we seek an \textit{HR cousin} that can serve as a reference image for RefSR of LR son. 

Note that any ZSSR methods including our RZSR can handle arbitrary scale factors~\cite{shocher2018zero, bell2019blind}. For brevity, we hereafter explain RZSR with a scale factor of 2. Let $I$, ${I^{ \downarrow 2}}$, and ${I^{ \downarrow 4}}$ be an input image and its downsampled images with scale factors of 2 and 4, respectively. Also let $P_{\bf{x}}$ denote an image patch of $I$ centered at position ${\bf{x}}$. $P_{\bf{x}}^{ \downarrow 2}$ and $P_{\bf{x}}^{ \downarrow 4}$ are defined similarly. If we define $P_{\bf{x}}^{ \downarrow 2}$ as LR son, then $P_{2\bf{x}}$ can be treated as its HR father. In cross-scale patch matching~\cite{glasner2009super}, the best match for LR son is found from ${I^{ \downarrow 4}}$. For example, if $P_{\bf{y}}^{ \downarrow 4}$ is the best match of $P_{\bf{x}}^{ \downarrow 2}$, then its father $P_{2\bf{y}}^{ \downarrow 2}$ is treated as an HR cousin. 

We attempt to obtain a high quality HR cousin without exhaustive cross-scale patch matching. Fig.~\ref{fig:patchmatch}(a) illustrates the HR cousin retrieval procedure in the training stage. First, the internal patch database, denoted as $\Theta^{ \downarrow 2}$, is generated from ${I^{ \downarrow 2}}$ using the method described in Section~\ref{sec:patchgen}. Then, instead of generating a new patch database from ${I^{ \downarrow 4}}$, we obtain directly $\Theta^{ \downarrow 4}$ from $\Theta^{ \downarrow 2}$ by simply 
extracting the corresponding patches from ${I^{ \downarrow 4}}$. Specifically, if ${\Theta ^{ \downarrow 2}}\, = \left\{ {P_{{{\bf{x}}_1}}^{ \downarrow 2},P_{{{\bf{x}}_2}}^{ \downarrow 2}, \cdot  \cdot  \cdot ,P_{{{\bf{x}}_N}}^{ \downarrow 2}} \right\}$, then ${\Theta ^{ \downarrow 4}} = \left\{ {P_{{{\bf{x}}_1}/2}^{ \downarrow 4},P_{{{\bf{x}}_2}/2}^{ \downarrow 4}, \cdot  \cdot  \cdot ,P_{{{\bf{x}}_N}/2}^{ \downarrow 4}} \right\}$. Let $d$ be the depth value at the center of $P_{\bf{x}}^{ \downarrow 2}$. Then, for every $P_{\bf{x}}^{ \downarrow 2}$, we retrieve its HR cousin using $\Theta^{ \downarrow 4}$ as follows:
\begin{equation}\label{eq:patch}
{{\bf{y}}^*} = \mathop {\arg \min }\limits_{{\bf{y}} \in {{\bf{Y}}^{ \downarrow 4}}} dist\left( {P_{\bf{x}}^{ \downarrow 2},P_{\bf{y}}^{ \downarrow 4}} \right),
\end{equation}
where ${{\bf{Y}}^{ \downarrow 4}}$ is a set of the center positions of the patches in $\Theta^{ \downarrow 4}$ with depth values smaller than $d$. Here, ${P_{\bf{x}}^{ \downarrow 2}}$ and ${P_{\bf{y}}^{ \downarrow 4}}$ have the same size and $dist$ measures their distance, where we used the VGG distance~\cite{johnson2016perceptual, zhang2019image}. Specifically, image-level VGG features are extracted from ${I^{ \downarrow 2}}$ and ${I^{ \downarrow 4}}$, respectively, and then the patch-level VGG features for ${P_{\bf{x}}^{ \downarrow 2}}$ and ${P_{\bf{y}}^{ \downarrow 4}}$ are extracted from the image-level VGG features using indexes ${\bf{x}}$ and ${\bf{y}}$. Finally, the HR cousin of $P_{\bf{x}}^{ \downarrow 2}$ is determined as $P_{2{\bf{y}}^*}^{ \downarrow 2}$.

Although we attempt to find the best HR cousin from the internal patch database, the HR cousin is not guaranteed to always help the SR of LR son. In particular, the HR cousin is not helpful when the depth map is unreliable or when the input image lacks cross-scale self-similarity. We thus retain the minimum distance while obtaining the best match using (\ref{eq:patch}) and compare it with a threshold $T$. If the minimum distance is greater than $T$, we use the bicubic upsampled version of $P_{\bf{x}}^{ \downarrow 2}$ as the HR cousin such that undesirable reference images do not harm the training of RZSR.

\begin{figure}[t] 
\begin{center}
\includegraphics[width=\linewidth,page=1]{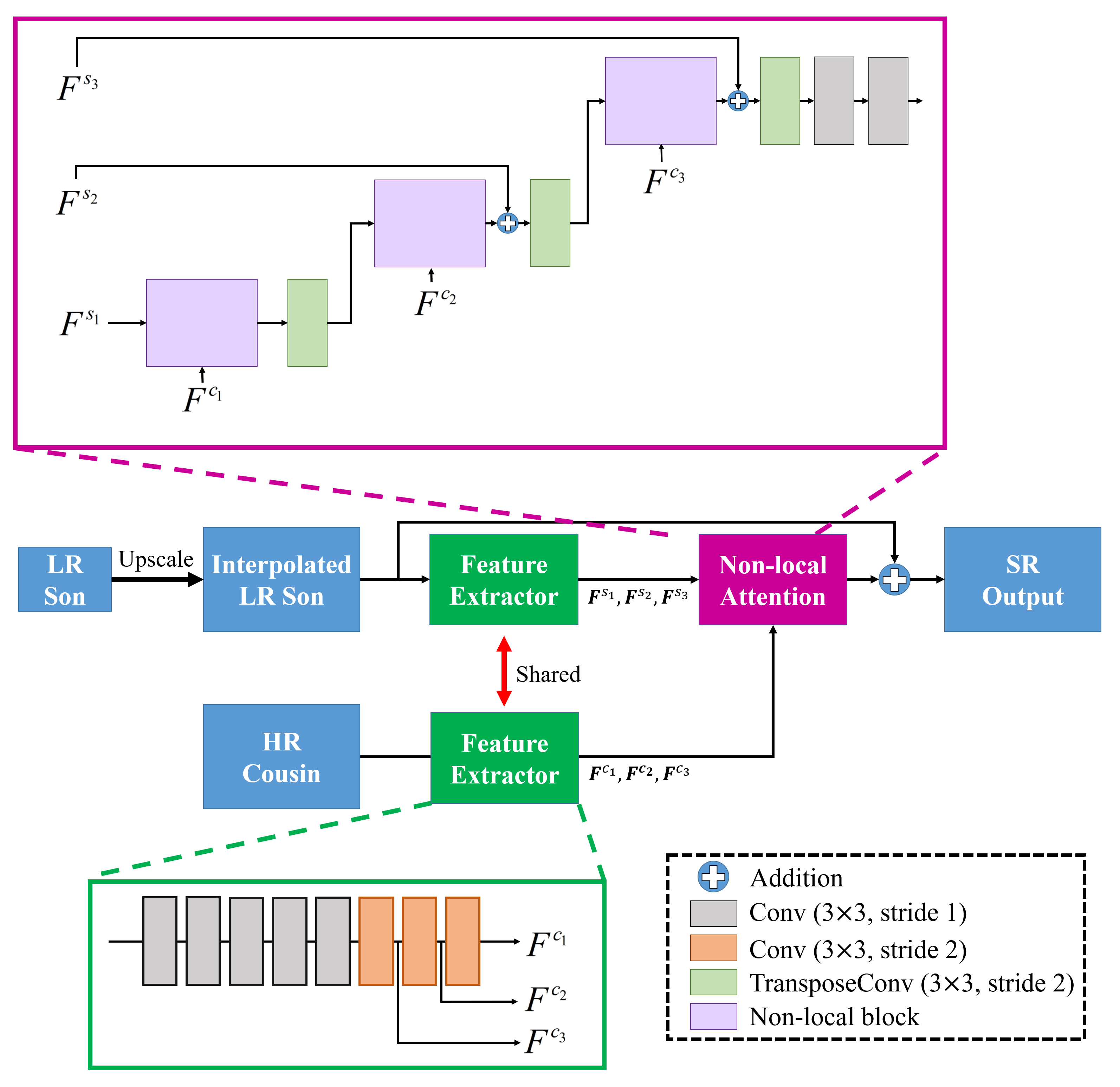}
\end{center}
   \caption{Illustration of the network architecture.}\label{fig:Network}
\end{figure}

\subsection{SR Network} \label{sec:network}
Given a triplet of LR son, HR father, and HR cousin, we train our SR network in a supervised manner. The overall architecture of the SR network is shown in Fig.~\ref{fig:Network}. Similar to \cite{shocher2018zero, bell2019blind}, LR son is upscaled to the output size and inputted to the network. The upscaled image is also added to the network output such that the network can only learn the residual. The shared feature extractor consists of eight convolutional layers with 128 channels, and the last three convolution layers are used with stride 2.

Our SR network is non-local in two aspects. First, since HR cousin is obtained from the internal patch database, its practical receptive field covers a whole image ideally. Second, to fully exploit the rich textures and high-frequency details of HR cousin, we use a non-local attention module inspired by \cite{wang2018non}.

\begin{figure}[htb]
    \centering
    \includegraphics[width=.45\textwidth]{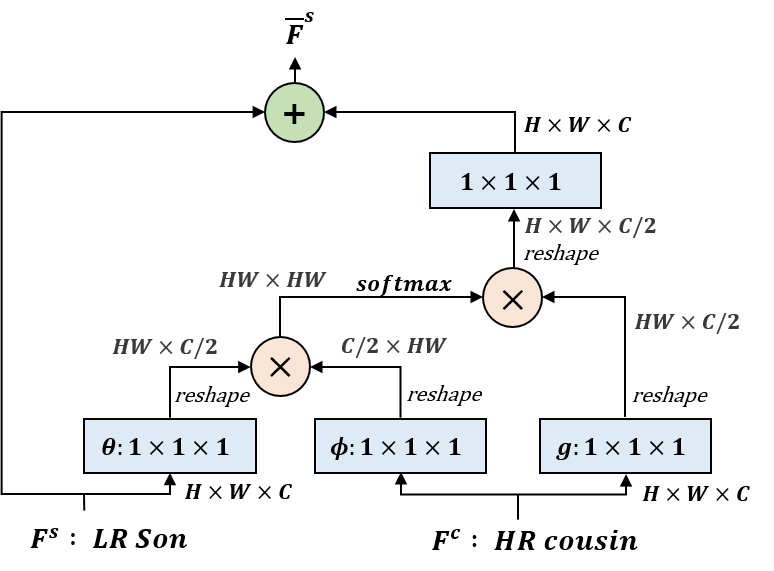}
    \caption{The structure of the non-local block.}
    \label{fig:nonlocal}
\end{figure}

Let ${F^s}$ and ${F^c}$ denote the features of LR son and HR cousin, respectively. Then, our non-local block as shown in Fig.~\ref{fig:nonlocal} is operated as follows:
\begin{equation}\label{eq:nonlocal}
\overline F _{\bf{x}}^s = F_{\bf{x}}^s + h\left( {\frac{1}{N}\sum\limits_{\forall {\bf{y}}} {\exp \left( {\theta {{\left( {F_{\bf{x}}^s} \right)}^T}\phi \left( {F_{\bf{y}}^c} \right)} \right)g\left( {F_{\bf{y}}^c} \right)} } \right),
\end{equation}
where $h$, $\theta$, $\phi$, and $g$ are 1$\times$1 convolutional layers for feature embedding, and $N$ is a normalization factor. ${\bf{x}}$ is the index of an output position, and ${\bf{y}}$ is the index that enumerates all positions. $F_{\bf{x}}^s$ represents the feature vector of $F^s$ at ${\bf{x}}$, and $\overline F{^s}$ is the output feature of the non-local block.

The operation in (\ref{eq:nonlocal}) is the same as the non-local block with the residual connection and embedded Gaussian in \cite{wang2018non} except that both ${F^s}$ and ${F^c}$ are used to obtain the output feature. Through this method, only HR features with high affinity to LR features can be added, resulting in a robust feature transfer from HR cousin to LR son. Motivated by \cite{shim2020robust}, our non-local attention module is embodied with multi-scale non-local blocks. To this end, three features with different scales are extracted from LR son and HR cousin, each of which is denoted as $F^{s_{1}}$, $\cdots$, $F^{c_{3}}$ in Fig.~\ref{fig:Network}. At the first scale, the non-local block is applied as described in (\ref{eq:nonlocal}) using $F^{s_{1}}$ and $F^{c_{1}}$. At the second scale, the output of the first non-local block is first resized using a transposed convolutional layer with stride 2 and then used as an input in place of $F^{s_{2}}$ for the second non-local block. The third scale is applied similarly. In this way, we can take advantage of multi-scale features from HR cousin. L2 loss is used to train our SR network.

\subsection{Inference} \label{sec:inference}
After learning how to upsample the downsampled image to the original image, actual SR is applied to the given image. Note that there is no HR father in the inference stage, and that LR son and HR cousin are obtained from the original image $I$. Fig.~\ref{fig:patchmatch}(b) illustrates the HR cousin retrieval procedure at the inference stage. For cross-scale matching of $I$, we use $\Theta^{ \downarrow 2}$, which has already been obtained at the training stage. With abuse of notation, let $d$ be the depth value at the center of $P_{\bf{x}}$. For each $P_{\bf{x}}$, we retrieve its HR cousin from $\Theta^{ \downarrow 2}$ as follows:
\begin{equation}\label{eq:patch-test}
{{\bf{y}}^*} = \mathop {\arg \min }\limits_{{\bf{y}} \in {{\bf{Y}}^{ \downarrow 2}}} dist\left( {{P_{\bf{x}}},P_{\bf{y}}^{ \downarrow 2}} \right),
\end{equation}
where ${{\bf{Y}}^{ \downarrow 2}}$ is a set of center positions of the patches in $\Theta^{ \downarrow 2}$ with depth values smaller than $d$. Consequently, the HR cousin of $P_{\bf{x}}$ is given as $P_{2{\bf{y}}^*}$. The same thresholding method described in Section~\ref{sec:patchretrieve} is also applied at the inference stage to avoid using an unreliable HR cousin. Note that patch matching is performed between ${P_{\bf{x}}^{ \downarrow 2}}$ and ${P_{\bf{y}}^{ \downarrow 4}}$ at training as (1), but between ${{P_{\bf{x}}}}$ and ${P_{\bf{y}}^{ \downarrow 2}}$ at inference as (3). 

Given LR son $P_{\bf{x}}$ and HR cousin $P_{2{\bf{y}}^*}$, an HR patch $P_{2{\bf{x}}}^*$ is obtained using the trained SR network, i.e., $P_{2{\bf{x}}}^* = {{\rm N}_\Phi }\left( {{P_{\bf{x}}},{P_{2{{\bf{y}}^*}}}} \right)$, where ${{\rm N}_\Phi }$ represents our SR network parameterized with $\Phi$. The same post-processing applied in the original ZSSR~\cite{shocher2018zero} is applied for further refinement.
For each $P_{\bf{x}}$ with size $M \times M$, we obtain $P_{2\bf{x}}^*$ with size $2M \times 2M$. The SR network is applied in a sliding window fashion with stride $s$, where we chose $s$ as 4, and the overlapped regions are averaged for full size image reconstruction.

\section{Experimental results}
\subsection{Dataset and Experimental Setup} \label{dataset}
For quantitative performance evaluation, we used the Set5~\cite{set5}, Set14~\cite{set14}, DIV2K~\cite{agustsson2017DIV2K}, and Urban100 datasets~\cite{huang2015single}. Following the traditional settings, LR images were first obtained by bicubic downsampling of the original images using the $imresize$ function in MATLAB~\cite{matlabresize}. For the test considering real-world environments, we used DIV2KRK~\cite{bell2019blind} and generated Set5RK, Set14RK, and Urban100RK by applying a random anisotropic Gaussian blur kernel to each image before downsampling, according to~\cite{bell2019blind}. To obtain a depth map for each test image, we used a pre-trained AdaBins model~\cite{bhat2021depth} for depth estimation. For the performance comparison with existing methods, we measured the PSNR and structural similarity (SSIM)~\cite{wang2004ssim} of the Y channel in the YCbCr color space.

We also evaluated the performance of SR methods on old photographs in the ``Historical'' dataset~\cite{historic}. Because the ground-truth HR images are not available for these images, we measured the no-reference image quality metrics, i.e., naturalness image quality evaluator (NIQE)~\cite{mittal2012making}, perceptual index (PI)~\cite{blau20182018}, and neural image assessment (NIMA) \cite{talebi2018}, from the resultant images.

We implemented the proposed network using PyTorch~\cite{pytorch} with a single Nvidia TITAN RTX GPU. We used Adam optimizer~\cite{kingma2014adam} with an initial learning rate of 0.001 and followed the learning rate decay rule of the original ZSSR~\cite{shocher2018zero}.
\begin{figure}[tp]
\begin{center}
\includegraphics[width=\linewidth,page=1]{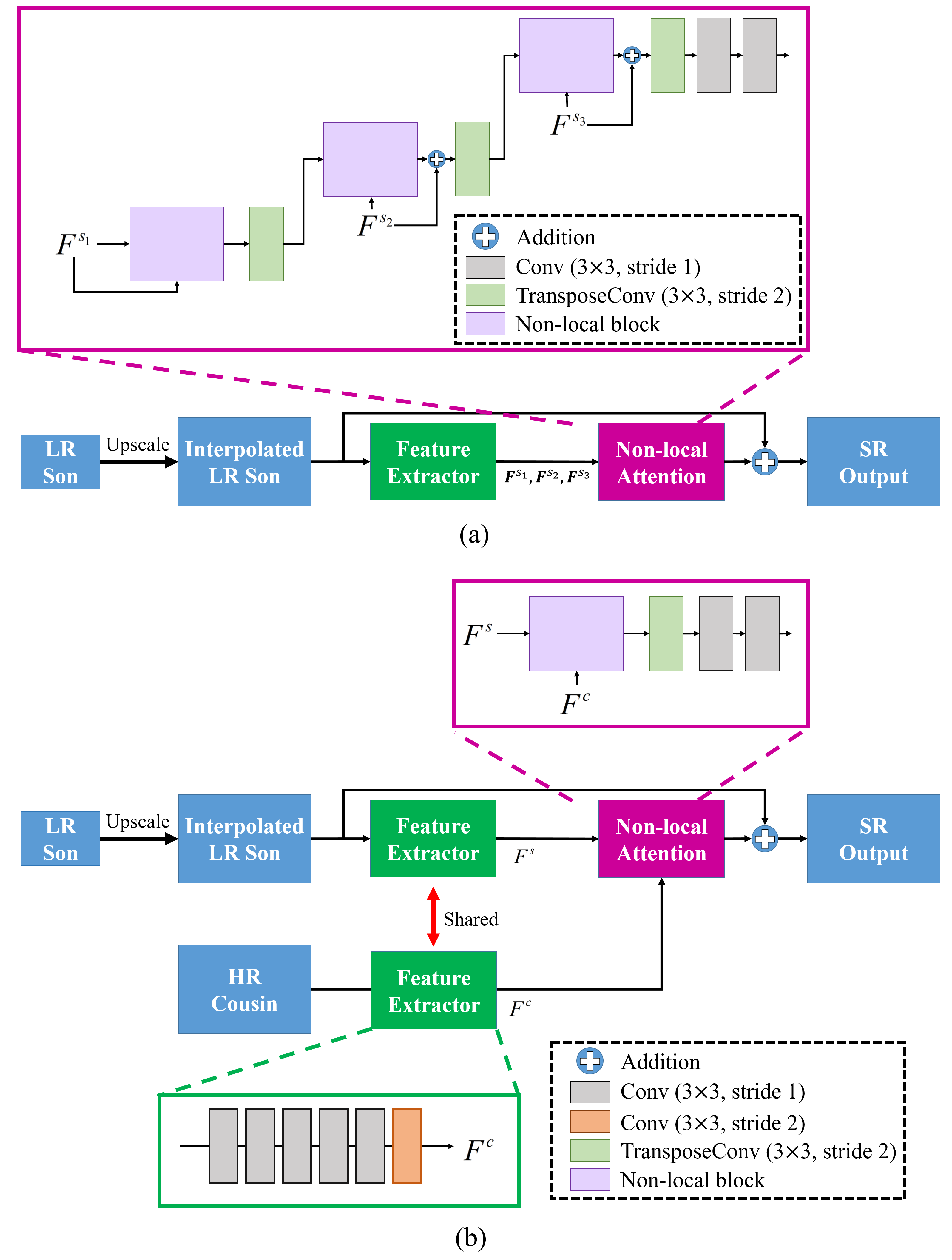}
\end{center}
   \caption{Models used for ablation studies: (a) SR network without the reference branch, (b) SR network with the single-scale non-local block.}\label{fig:Network-ablation}
\end{figure}

\subsection{Parameter Settings and Ablation Studies} \label{ablation}

We chose the default settings for the number of depth ranges $D$ and the threshold $T$ as 5 and 0.9, respectively. We varied the number of clusters $k$ for each depth range. Specifically, we set $k_i$ as $\left\lceil {{N_i}/100} \right\rceil $, where ${k_i}$ is the number of clusters in the $i$-th depth range, ${N_i}$ is the number of its corresponding image patches, and $\left\lceil \cdot \right\rceil $ is the ceiling function. 
Under the default settings, we varied each parameter individually to test the impact on parameter changes. For this test, DIV2K and Urban100 were used for $\times$ 2 SR. As shown in Table~\ref{table:ablation_1}, we obtained the best performance from the default settings for both datasets. 

\begin{table}[htb]
\centering
\caption{Performance evaluation of RZSR under different parameter settings for 2$\times$SR of Urban100 and DIV2K in terms of PSNR (dB) / SSIM.  }
\begin{tabular}{|c|c|c|c|}
	\hline
 \multicolumn{2}{|c|}{Parameter} & Urban100 & DIV2K \\ \hline
\multirow{3}{*}{$T$} 
& 1.0 & 30.75 / 0.9130 & 34.90 / 0.9429 \\
& 0.9 & 30.77 / 0.9140 & 34.94 / 0.9438 \\
& 0.8 & 30.70 / 0.9119 & 34.86 / 0.9419 \\ 
& 0.7 & 30.68 / 0.9110 & 34.73 / 0.9401 \\ \hline
\multirow{3}{*}{$D$} & 3 & 30.69 / 0.9120   & 34.88 / 0.9428 \\
& 4 & 30.74 / 0.9133 & 34.92 / 0.9433 \\ 
& 5 & 30.77 / 0.9140 & 34.94 / 0.9438 \\ 
& 6 & 30.67 / 0.9115 & 34.83 / 0.9420 \\ \hline 
\end{tabular}
\label{table:ablation_1}
\end{table}
\begin{table}[htb]
\centering
\caption{Ablation studies of RZSR for 2$\times$SR of Urban100 and DIV2K. The performance is measured as PSNR (dB) / processing time (s). ES: exhaustive search.}
\begin{tabular}{|c|c|c|c|}
	\hline
Patch size & Model & Urban100& DIV2K \\  \hline
\multirow{4}{*}{32$\times$32} &  Fig.~\ref{fig:Network-ablation}(a) & 30.08 / 160.48 & 34.47 / 332.21 \\
& Fig.~\ref{fig:Network-ablation}(b) & 30.11 / 194.62  & 34.56 / 369.46 \\
& Fig.~\ref{fig:Network} w/ ES & 30.49 / 312.47  &  34.86 / 556.10\\ 
& Fig.~\ref{fig:Network} & 30.53 / 270.97  &  34.80 / 492.49\\ \hline
\multirow{4}{*}{48$\times$48} & Fig.~\ref{fig:Network-ablation}(a) & 30.29 / 207.82 &  34.52 / 340.97 \\
& Fig.~\ref{fig:Network-ablation}(b) & 30.19 / 242.17  & 34.67 / 402.73 \\
& Fig.~\ref{fig:Network} w/ ES &  30.74 / 424.78  & 34.93 / 609.32\\ 
& Fig.~\ref{fig:Network}  & 30.77 / 372.24  &  34.94 / 509.42 \\ \hline

\multirow{4}{*}{64$\times$64} & Fig.~\ref{fig:Network-ablation}(a) & 30.22 / 277.41   & 34.50 / 401.89 \\
&  Fig.~\ref{fig:Network-ablation}(b) & 30.28 / 319.77 & 34.69 / 476.44\\
& Fig.~\ref{fig:Network} w/ ES & 30.74 / 480.12  & 34.96 / 645.89\\ 
& Fig.~\ref{fig:Network}  & 30.75 / 420.53    & 34.91 / 570.16  \\ \hline

\end{tabular}
\label{table:ablation_2}
\end{table}

\begin{comment}
\begin{table}[htb]
\centering
\caption{Ablation studies of RZSR for 2$\times$SR of Urban100 and DIV2K. The performance is measured as PSNR (dB) / processing time (s).}
\begin{tabular}{|c|c|c|c|c|}
	\hline
 \multirow{2}{*}{\!\!Patch size\!\!}& Patch & Reference& \multirow{2}{*}{Urban100}& \multirow{2}{*}{DIV2K} \\ 
 & database & branch & &\\  \hline
\multirow{3}{*}{32$\times$32} & &  & 30.08 / 160.48 & 34.47 / 332.21 \\
& &\checkmark  & 30.49 / 312.47  &  34.86 / 556.10\\ 
&\checkmark &\checkmark  &30.53 / 270.97  &  34.80 / 492.49\\ \hline
\multirow{3}{*}{48$\times$48} & &  & 30.29 / 207.82 &  34.52 / 340.97 \\
& &\checkmark  &  30.74 / 424.78  & 34.93 / 609.32\\ 
&\checkmark &\checkmark  & 30.77 / 372.24  &  34.94 / 509.42 \\ \hline
\multirow{3}{*}{64$\times$64} & &  & 30.22 / 277.41   & 34.50 / 401.89 \\
& &\checkmark  & 30.74 / 480.12  & 34.96 / 645.89\\ 
&\checkmark &\checkmark  & 30.75 / 420.53    & 34.91 / 570.16  \\ \hline

\end{tabular}
\label{table:ablation_2}
\end{table}
\end{comment}

\begin{table}[htb]
\centering
\caption{Performance evaluation for different depth estimation methods. The performance is measured as PSNR (dB) / SSIM / processing time (s) for 2$\times$SR of bicubic downsampled Urban100 and DIV2K images. }
\begin{tabular}{|c|c|c|}
	\hline
Method & Urban100 & DIV2K \\ \hline
w/o depth & 30.74 / 0.9127 / 424.78 & 34.93 / 0.9440 / 609.32 \\ \hline
Monodepth2~\cite{godard2019digging} & 30.71 / 0.9122 / 365.09 & 34.92 / 0.9437 / 498.17 \\ \hline
AdaBins~\cite{bhat2021depth} & 30.77 / 0.9139 / 372.24 & 34.94 / 0.9439 / 509.42 \\ \hline

\end{tabular}
\label{table:Depth}
\end{table}

\begin{table*}[htb]
\centering
\caption{Full-reference quality evaluation (PSNR/SSIM) of the SR methods on the bicubic-downsampled images. For comparison with other works, the PSNR and SSIM values of the Y channel are reported. The best and second-best scores are boldfaced and underlined, respectively. $^{\dag}$These networks were trained using external datasets.}
\begin{tabular}{|l|c|c|c|c|c|}
	\hline
	Method & Scale & Set5 & Set14 & Urban100 & DIV2K \\ \hline
	\multirow{3}{*}{Bicubic interpolation} & 2$\times$ & 33.66 / 0.9299 & 30.24 / 0.8688 &  26.87 / 0.8496 & 32.41 / 0.9110 \\ 
	& 3$\times$ & 30.39 / 0.8682 & 27.55 / 0.7742 &  24.47 / 0.7349 & 29.64 / 0.8423 \\ 
	& 4$\times$ & 28.42 / 0.8104 & 26.00 / 0.7027 &  23.14 / 0.6577 & 28.10 / 0.7850 \\ \hline 
    
    \multirow{3}{*}{$^{\dag}$EDSR+\cite{lim2017enhanced}} & 2$\times$ &38.20 / 0.9606 & 34.02 / 0.9204 & 33.10 / 0.9363  & 36.06 / 0.9497\\
    & 3$\times$ & \underline{34.76} / 0.9209 & \underline{30.66} / 0.8481 & 29.02 / 0.8685  & 32.37 / 0.8975\\
    & 4$\times$ & 32.62 / 0.8984 & 28.94 / \underline{0.7901} & 26.89 / 0.8080  & 30.41 / 0.8487\\ \hline
    
    \multirow{3}{*}{$^{\dag}$SAN+\cite{dai2019second}} & 2$\times$ & \textbf{38.35} / \textbf{0.9619} & \textbf{34.44} / \textbf{0.9244} & \textbf{33.73} / \textbf{0.9416} & 36.61 / 0.9533\\ 
    & 3$\times$ & \textbf{34.89} / \textbf{0.9306} & \textbf{30.77} / \textbf{0.8498} & \textbf{29.29} / \textbf{0.8730} & \underline{32.95} / \underline{0.9048}\\ 
    & 4$\times$ & \textbf{32.70} / \textbf{0.9013} & \textbf{29.05} / \textbf{0.7921} & \textbf{27.23} / \textbf{0.8169} & \underline{31.08} / \underline{0.8614}\\ \hline
    \multirow{3}{*}{$^{\dag}$CSNLN~\cite{mei2020image}} &2$\times$ & \underline{38.28} / \underline{0.9616} & \underline{34.12} / \underline{0.9223}  & 33.25 / 0.9386 & \textbf{36.79} / \textbf{0.9547}\\
    &3$\times$ & 34.74 / \underline{0.9300} &\underline{30.66} / 0.8482  & 29.13 / 0.8712 & 32.81 / 0.9036\\
    &4$\times$ & \underline{32.68} / \underline{0.9004} & \underline{28.95} / 0.7888 & \underline{27.22} / \underline{0.8168}& 30.88 / 0.8583\\ \hline
    \multirow{3}{*}{$^{\dag}$IGNN~\cite{zhou2020crossscale}} &$2\times$& 38.24 / 0.9613 & 34.07 / 0.9217 & \underline{33.42} / \underline{0.9396} & \underline{36.74} / \underline{0.9540}\\
    &3$\times$& 34.72 / 0.9298 & \underline{30.66} / \underline{0.8484} & \underline{29.20} / \underline{0.8721} & \textbf{33.06} / \textbf{0.9060}\\
    &4$\times$& 32.57 / 0.8998 & 28.85 / 0.7891& 27.04 / 0.8128 & \textbf{31.09} / \textbf{0.8615}\\ \hline
    \multirow{3}{*}{ZSSR~\cite{shocher2018zero}} &2$\times$& 36.93 / 0.9570 & 32.67 / 0.9108 & 29.54 / 0.9034 & 34.39 / 0.9379\\
    &3$\times$& 32.05 / 0.9073& 29.04 / 0.8350 & 25.94 / 0.8051 & 30.91 / 0.8745\\
    &4$\times$& 29.04 / 0.8352 &27.00 / 0.7618 & 24.13 / 0.7232 & 29.08 / 0.8183\\ \hline
    \multirow{3}{*}{RZSR (ours)} & 2$\times$& 37.70 / 0.9600  & 33.49 / 0.9160  & 30.77 / 0.9140 & 34.94 / 0.9438 \\
    &3$\times$& 34.01 / 0.9209 & 30.16 / 0.8411 & 27.47 / 0.8598 & 31.38 / 0.8907\\
    &4$\times$& 31.82 / 0.8864 & 28.31 / 0.7760 & 25.29 / 0.7742 & 29.42 / 0.8310\\
    \hline

\end{tabular}
\label{table:full-res-score_Bicubic}
\end{table*}

\begin{table*}[htb!]
\centering
\caption{Full-reference quality evaluation (PSNR/SSIM) of the SR methods on the images downsampled with random kernels. For comparison with other works, the PSNR and SSIM values of the Y channel are reported. The best and second-best scores are boldfaced and underlined, respectively. $^{\dag}$These networks were trained using external datasets.}
\begin{tabular}{|c|l|c|c|c|c|c|}
	\hline
	Category & Method & Scale & Set5RK & Set14RK & Urban100RK & DIV2KRK \\ \hline
	\multirow{21}{*}{Bicubically trained}  & \multirow{3}{*}{Bicubic interpolation} & 2$\times$ & 28.50 / 0.8463 & 26.29 / 0.7395 & 23.42 / 0.7055 & 28.73 / 0.8040 \\ 
    & & 3$\times$ & 27.76 / 0.8153 & 25.47 / 0.7054 & 22.78 / 0.6625 & 27.44 / 0.7686 \\
    & & 4$\times$ & 26.74 / 0.7698 & 25.00 / 0.6733 & 22.27 / 0.6224 & 26.94 / 0.7434 \\ \cline{2-7}
    
    & \multirow{3}{*}{$^{\dag}$EDSR+\cite{lim2017enhanced}} &2$\times$ & 28.88 / 0.8598 & 26.58 / 0.7532 & 23.01 / 0.6857  & 29.17 / 0.8216\\
    &&3$\times$ & 28.51 / 0.8456 & \underline{26.10} / 0.7378 & 23.43 / 0.7073  & 28.01 / 0.7940\\
    &&4$\times$ & 27.83 / 0.8180 & 26.03 / 0.7251 & \underline{23.16} / 0.6879  & 27.77 / 0.7817\\ \cline{2-7}
    
    & \multirow{3}{*}{$^{\dag}$SAN+\cite{dai2019second}} & 2$\times$ & \underline{28.92} / \underline{0.8606} & \underline{26.60} / \underline{0.7540} & \underline{23.80} / \underline{0.7289} & \underline{29.21} /0.8232 \\ 
    & & 3$\times$ & 28.50 / 0.8461 & 26.03 / 0.7391 & \underline{23.44} / 0.7097 & 28.04 / 0.7954 \\
    & & 4$\times$ & 27.99 / 0.8208 & \underline{26.08} / \underline{0.7288} & 23.14 / 0.6916 & 27.81 / \textbf{0.7851} \\ \cline{2-7}
    & \multirow{3}{*}{$^{\dag}$CSNLN~\cite{mei2020image}} & 2$\times$ & 28.89 / 0.8602 & 26.59 / 0.7536 & 23.31 / 0.6977 & 29.20 / \textbf{0.8313} \\ 
    & & 3$\times$ & \underline{28.54} / \underline{0.8467} & 26.04 / \underline{0.7399} & \underline{23.44} / \underline{0.7099} & 28.04 / 0.7954 \\
    & & 4$\times$ & \underline{28.00} / 0.8207 & \textbf{26.10} / 0.7285 & 23.14 / \underline{0.6918} & \underline{27.82} / \underline{0.7847} \\ \cline{2-7}
    & \multirow{3}{*}{$^{\dag}$IGNN~\cite{zhou2020crossscale}} & 2$\times$ & 28.78 / 0.8604 & 26.55 / 0.7539 & 23.32 / 0.6977 & 29.20 /  \underline{0.8310} \\ 
    & & 3$\times$ & 28.53 / 0.8456 & 26.02 / 0.7389 &\underline{23.44} / 0.7096 & \underline{28.05} / \underline{0.7955} \\
    & & 4$\times$ & \textbf{28.01} / \textbf{0.8211} & 26.05 / 0.7282 & 23.12 / 0.6908 & 27.80 / \underline{0.7847} \\ \cline{2-7}
    & \multirow{3}{*}{ZSSR~\cite{shocher2018zero}} & 2$\times$ & 28.87 / 0.8601 & 26.57 / \underline{0.7540} & 23.72 / 0.7254 & 29.10 / 0.8215 \\ 
    & & 3$\times$ & 28.38 / 0.8398 & 25.88 / 0.7330 & 23.29 / 0.6964 & 27.87 / 0.7891 \\
    & & 4$\times$ & 27.29 / 0.7909 & 25.70 / 0.7085 & 22.85 / 0.6610 & 27.53 / 0.7708 \\ \cline{2-7}
    & \multirow{3}{*}{RZSR (ours)} & 2$\times$ & \textbf{29.00} / \textbf{0.8622} & \textbf{26.73} / \textbf{0.7546} & \textbf{23.88} / \textbf{0.7314} & \textbf{29.23} / 0.8230\\ 
    & & 3$\times$ & \textbf{28.66} / \textbf{0.8497} & \textbf{26.22} / \textbf{0.7418} & \textbf{23.52} / \textbf{0.7120} & \textbf{28.09} / \textbf{0.7960} \\
    & & 4$\times$ & 27.98 / \underline{0.8210} & \textbf{26.10} / \textbf{0.7291} & \textbf{23.20} / \textbf{0.6919} & \textbf{27.83} / 0.7831 \\ \hline \hline
    \multirow{5}{*}{Blind}  
    & BlindSR~\cite{cornillere2019sr} & \multirow{5}{*}{2$\times$} & 26.29 / 0.8203 & 24.72 / 0.7219 & 22.17 / 0.6716 & 29.41 / 0.8546\\ 
    & DualSR~\cite{emad2021} & & 28.76 / 0.8612 & \underline{27.10} / \underline{0.7977} & \underline{25.04} / \underline{0.7803} & \textbf{30.92} / \textbf{0.8728}\\ 
    & IKC~\cite{gu2019sr} &  & 28.17 / 0.8253 & 25.96 / 0.7084 & 22.87 / 0.6604 & 28.11 / 0.7809\\  
    & ZSSR + KernelGAN~\cite{bell2019blind}& &\underline{29.60} / \underline{0.8863}  & 26.73 / 0.7938 & 24.67 / 0.7706 & 30.36 / 0.8669 \\  
    & RZSR (ours) + KernelGAN & & \textbf{32.19} / \textbf{0.9135} & \textbf{28.27} / \textbf{0.8209} & \textbf{25.84} / \textbf{0.7929} & \underline{30.55} / \underline{0.8704}   \\ \hline 

\end{tabular}
\label{table:full-res-score}
\end{table*}

\begin{figure}[htb]
\begin{center}
\includegraphics[width=\linewidth,page=1]{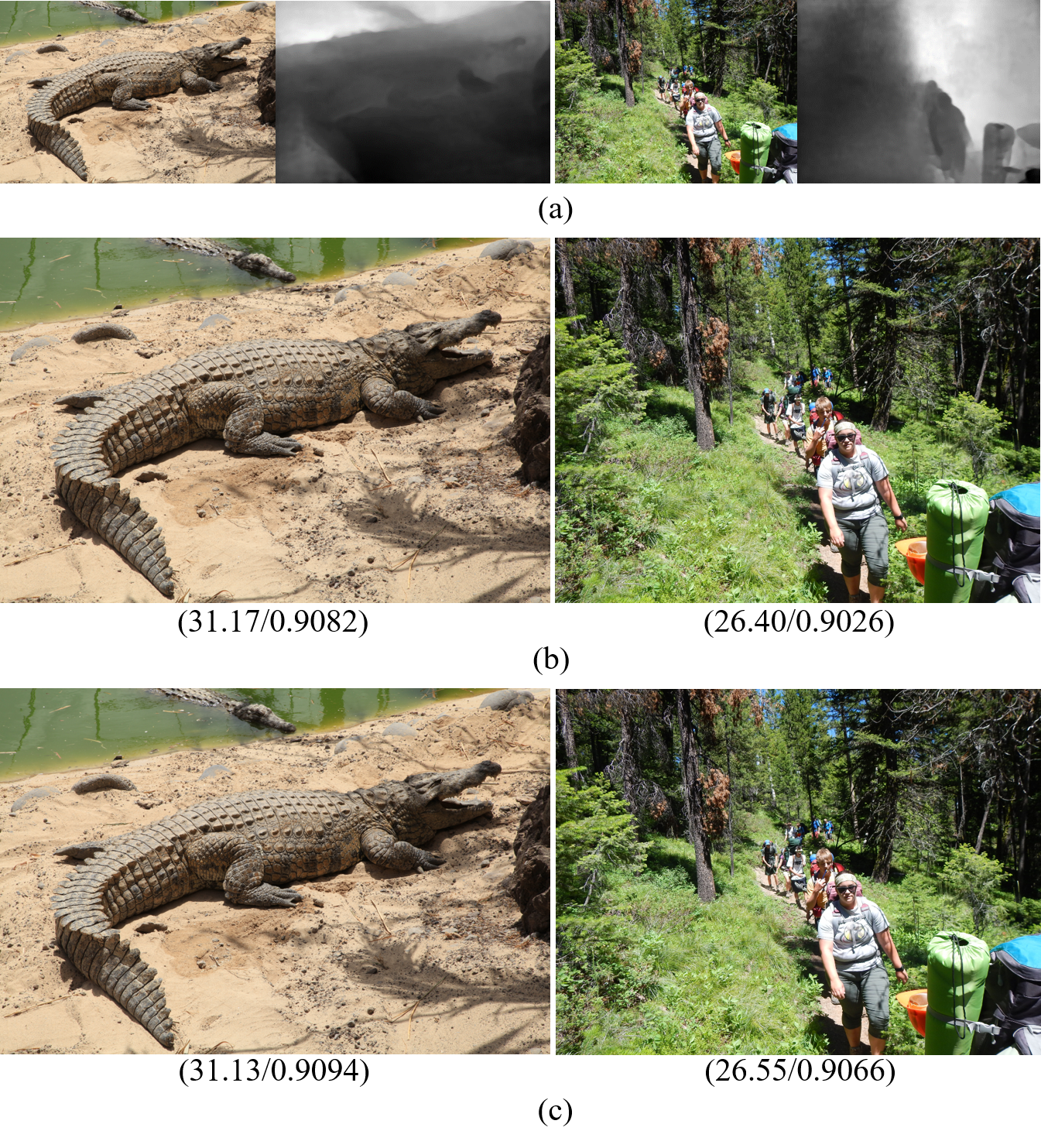}
\end{center}
   \caption{Failure cases: (a) LR input images and depth maps obtained using AdaBins~\cite{bhat2021depth}, (b) SR results obtained using the reference-free model shown in Fig.~\ref{fig:Network-ablation}(a), and (c) SR results obtained by the proposed method. The quality scores (PSNR/SSIM) are provided below each result.}\label{fig:failure}
\end{figure}

As the patch size $M \times M$ increases, the number of patches applicable for training decreases. The number of elements in the reference patch dataset also decreases as $M$ increases, and it thus becomes less likely that reference images can help the SR of unnecessarily large input LR patches. However, SR networks generally prefer using a large receptive field to learn sufficient features. To this end, we compared the performance using different patch sizes, i.e., $32 \times 32$, $48 \times 48$, and $64 \times 64$. Based on the result listed in Table~\ref{table:ablation_2}, we chose the default patch size as $48 \times 48$.                    

\begin{figure*}[htp]
\begin{center}
\includegraphics[width=0.98\linewidth]{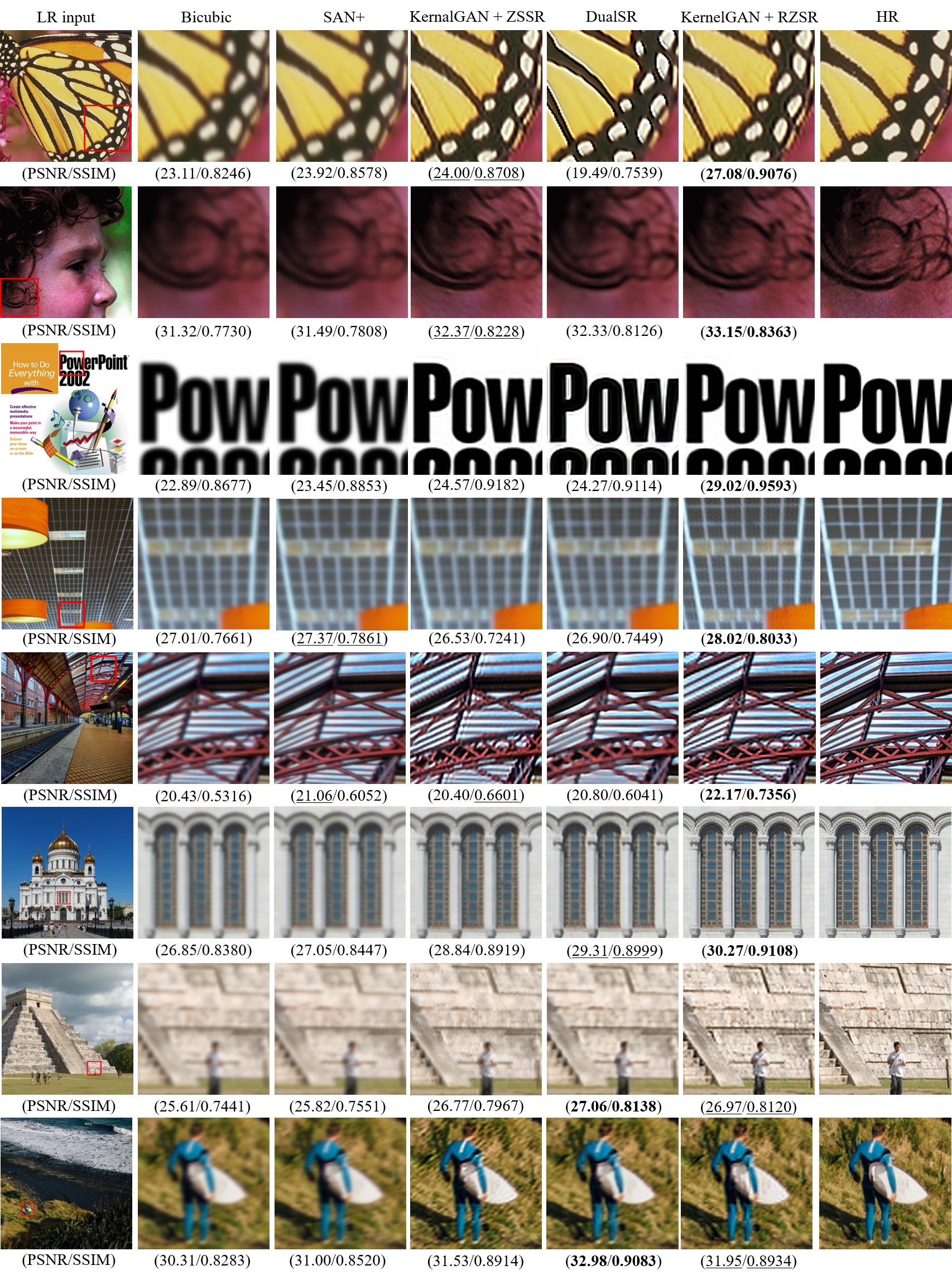}
\end{center}
   \caption{Visual quality comparison on several images of Set5RK (first two rows), Set14RK (third row), Urban100RK (fourth and fifth rows) and DIV2KRK (last three rows) for $\times$2 SR. }\label{fig:results-div-urban}
\end{figure*}

\begin{figure*}[htb!] 
\begin{center}
\includegraphics[width=0.98\linewidth]{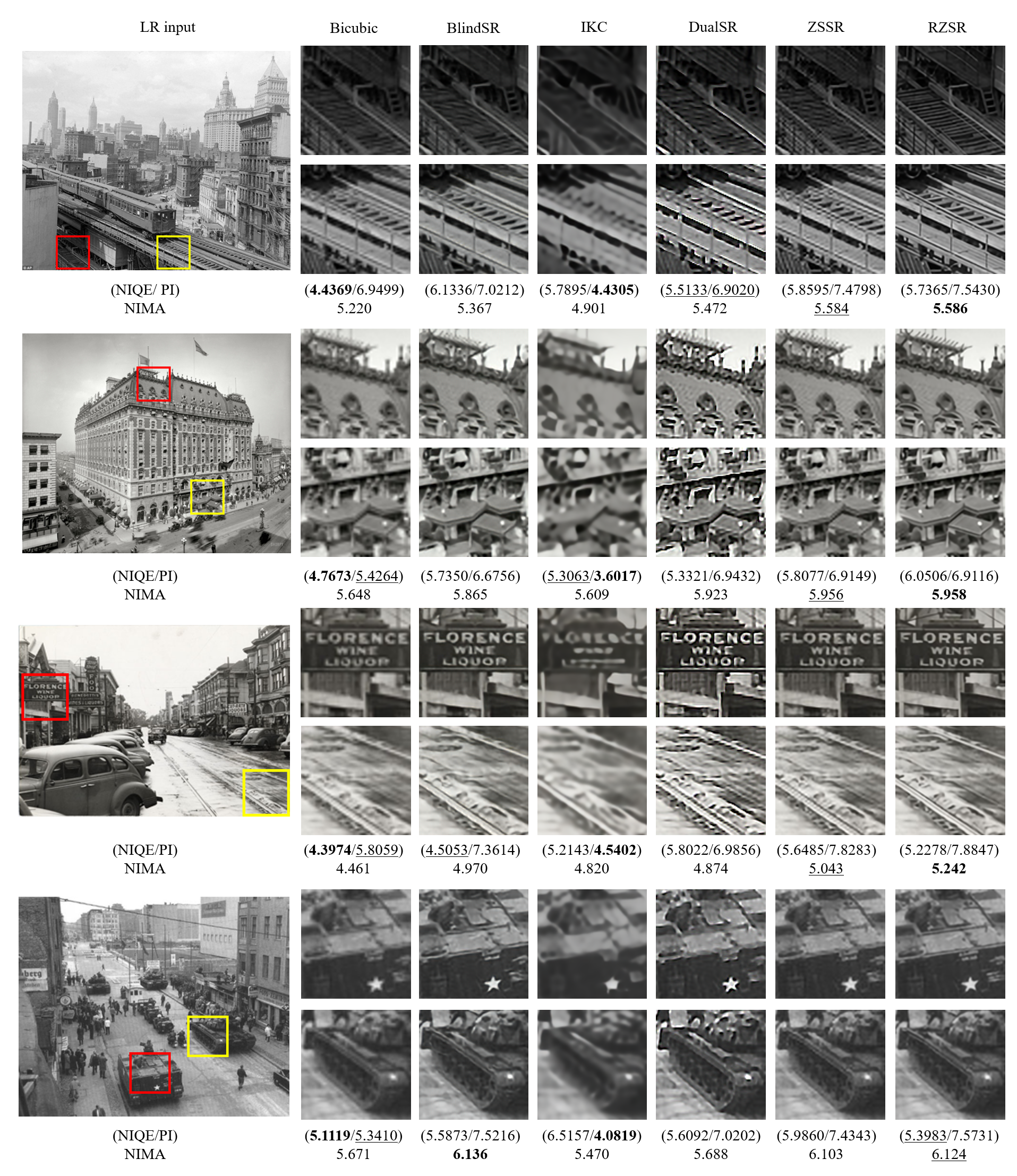}
\end{center}
   \caption{Results on the real-world images in the Historical dataset with no-reference quality evaluation scores (NIQE/PI/NIMA) for $\times$2 SR.}\label{fig:old_photo}
\end{figure*}

To evaluate the performance improvement obtained using the reference images in RZSR, we tested its reference-free version, as shown in Fig.~\ref{fig:Network-ablation}(a). Note that the network structure is similar to Fig.~\ref{fig:Network} except that the non-local attention module performs self-attention using multi-scale non-local blocks. Although the non-local attention module itself can exploit the self-similarity of input images, the performance drops from RZSR were obtained as 0.48 dB and 0.42 dB for Urban100 and DIV2K, respectively, for the patch size $48 \times 48$.

Next, to show the effectiveness of the multi-scale non-local blocks, we tested the model embodied with a single-scale non-local block, as shown in Fig.~\ref{fig:Network-ablation}(b). The use of the single-scale non-local block required less processing time (130.07s and 106.69s faster for Urban100 and DIV2K, respectively); however, the performance drops from RZSR were also not negligible (0.58 dB and 0.27 dB lower for Urban100 and DIV2K, respectively) for the patch size of $48 \times 48$. 

We also compared the performance of RZSR with and without the use of the internal patch database to investigate its effectiveness. In the case without the internal patch, we obtained HR cousins through an exhaustive search without using clustering or depth information (w/ ES in Table~\ref{table:ablation_2}). The proposed method of retrieving reference images using depth information and clustering resulted in comparable performance (0.03 dB and 0.01 dB higher for Urban100 and DIV2K, respectively) but required much less computational cost than the method with the exhaustive search (53s and 100s faster for Urban100 and DIV2K, respectively) for the patch size of $48 \times 48$. 

Last, we investigated the influence of the depth map on the performance of RZSR. To this end, we compared the performance of RZSR obtained using the state-of-the-art AdaBins~\cite{bhat2021depth} and widely-used Monodepth2 models~\cite{godard2019digging}. In the case without using the depth map, an exhaustive search was performed to obtain HR cousins. Compared to this case, both monocular depth estimation models contributed similarly to the performance by providing similar PSNR and SSIM scores but decreasing the processing time significantly, as presented in Table~\ref{table:Depth}.
The state-of-the-art AdaBins model still failed to provide reliable depth maps for some images, as shown in Fig.~\ref{fig:failure}. In such cases, unreliable HR cousins are replaced by bicubic upsampled versions of LR sons according to the thresholding method described in Sec.~\ref{sec:patchretrieve}. Consequently, the proposed method performed comparably to its reference-free version for these failure cases.

\subsection{Performance Comparison} \label{eval-quantitative}
Table~\ref{table:full-res-score_Bicubic} reports the PSNR and SSIM scores of our RZSR and the other SR methods on the bicubic-downsampled images. Under this traditional but impractical setting, the fully-supervised SR networks, especially SAN+ \cite{dai2019second}, outperformed ZSSR~\cite{shocher2018zero} and RZSR by large margins for all scale factors. The proposed RZSR still produced significantly higher PSNRs than ZSSR, e.g., 0.77 dB, 0.82 dB, 1.23 dB, and 0.55 dB improvements for $\times$2 SR of the Set5, Set14, Urban100, and DIV2K datasets, respectively.

Table~\ref{table:full-res-score} provides the PSNR and SSIM scores on the downsampled images obtained with random kernels. Here, the SR methods are divided into two categories, where the first category ``bicubically trained'' represents SR methods trained on bicubic downsampled images. When the bicubically trained SR models, i.e., EDSR+\cite{lim2017enhanced}, SAN+\cite{dai2019second}, CSNLN~\cite{mei2020image}, and IGNN~\cite{zhou2020crossscale}, were tested on Set5RK, Set14RK, DIV2KRK and Urban100RK, the performance dropped significantly, indicating their vulnerability to real-world LR images. Among the bicubically trained ones, RZSR showed the highest PSNR scores for all scale factors and all datasets, except for 4$\times$ SR on Set5RK.

The second category ``blind'' refers to SR methods that estimate downsampling kernels for robust SR. We found these blind SR methods, i.e., BlindSR~\cite{cornillere2019sr}, DualSR~\cite{emad2021}, and IKC~\cite{gu2019sr}, performed unstably for large scale factors, and thus the performance comparison was conducted using only the scale factor of 2. For ZSSR~\cite{shocher2018zero} and RZSR, we estimated kernels using KernelGAN~\cite{bell2019blind}, and we used the estimated kernels when downsampling the input images for training, whereas DualSR~\cite{emad2021} and IKC~\cite{gu2019sr} estimated kernels during SR. These blind methods, especially DualSR, ZSSR with KernelGAN, and RZSR with KernelGAN, resulted in higher PSNR and SSIM scores than the bicubically trained methods. RZSR with KernelGAN produced 0.59 dB, 1.17 dB, and 0.80 dB higher PSNR scores than the second-best method for Set5RK, Set14RK, and Urban100RK, respectively, and the second-best PSRN score for DIV2KRK.

\begin{figure}[t]
\begin{center}
\includegraphics[width=0.5\textwidth,page=1]{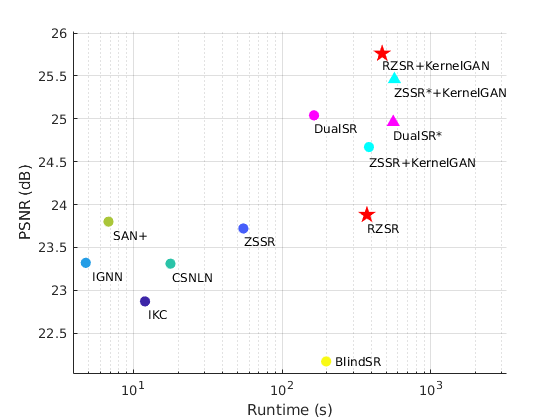}
\end{center}
  \caption{Comparison between the proposed and conventional methods in terms of the PSNR and runtime for $\times$2 SR of the Urban100RK dataset.}\label{fig:tradeoff}
\end{figure}

\begin{table}[t]
\centering
\caption{Performance comparison of SR methods in terms of NIQE, PI, NIMA scores for the Historical dataset. }
\begin{tabular}{|c|c|c|c|}
	\hline
 Method & NIQE & PI & NIMA \\ \hline
 
Bicubic & \textbf{4.9190} & 4.8245 & 5.2776  \\ \hline  
IKC & 5.6538 & 5.7962 & 5.3193  \\ \hline
DualSR & \underline{5.3855} & \underline{4.2379} & 5.4771  \\ \hline
BlindSR & 5.7864 & 4.5187 & 5.5598  \\ \hline
ZSSR & 5.8260 & 4.4468 & \underline{5.6296}  \\ \hline
RZSR & 5.4649 & \textbf{4.2337} & \textbf{5.6680}  \\ \hline
\end{tabular}
\label{table:NIMA_score}
\end{table}

Fig.~\ref{fig:results-div-urban} shows a visual comparison of the results. Overall, the proposed method produced sharper edges and finer details than the other methods. SAN+ did not show a clear advantage over bicubic upsampling for these LR images obtained with random kernels. Although DualSR exhibited higher PSNRs than RZSR with KernelGAN for the last two image samples, the structures and texture details are less vivid than those obtained by RZSR with KernelGAN. Fig.~\ref{fig:old_photo} further compares the results on challenging real-world LR images in the Historical dataset~\cite{historic}. Because original HR images are not available, NIQE, PI, and NIMA scores are provided below each result, and the average scores are listed in Table~\ref{table:NIMA_score}.
KernelGAN was found to be less robust to these old photographs and thus not applied with ZSSR and RZSR. The proposed RZSR renders HR images with fewer artifacts than the other methods and exhibits the lowest PI score (lower is better) and highest NIMA score (higher is better). We found that NIQE is not very closely correlated with subjective quality for our task. For example, the lowest (lower is better) NIQE was obtained from the simplest bicubic interpolation as reported in Table~\ref{table:NIMA_score}, but the resultant images obtained by bicubic interpolation are very blurry as shown in Fig.~\ref{fig:old_photo}.

Finally, Fig.~\ref{fig:tradeoff} compares the SR methods in terms of both PSNR and processing time for $\times$2 SR of the Urban100RK dataset. As can be seen, blind SR methods, including ZSSR, RZSR, DualSR, and BlindSR, required much longer processing time than the other SR methods developed for the non-blind setting. When we extended DualSR and ZSSR by increasing the number of layers in their architectures, denoted as DualSR* and ZSSR*, the resultant PSNRs were still significantly lower than the PSNR obtained by RZSR with KernelGAN. In other words, the highest performance of RZSR with KernelGAN cannot be simply obtained by augmenting DualSR or ZSSR with KernelGAN. High computational demand is still a drawback of RZSR, which will be addressed in future work. More results and source code are available in our project page\footnote{https://github.com/junsang7777/RZSR}.

\section{Conclusion}

In this paper, we proposed RZSR that integrates RefSR and ZSSR. By constructing an image-specific internal reference dataset using clustering and depth information, we were able to obtain reference image patches with rich textures and high-frequency details. Moreover, a RefSR network embodied with a non-local attention module was developed to more aggressively exploit image self-similarity. Experimental results demonstrated the effectiveness of RZSR.
Several future research directions can be considered based on the limitations of this study. First, to remain faithful to the ZSL principle, we did not pretrain our SR network using external images. There is another research direction that allows network pretraining to ZSL, called generalized ZSL~\cite{verma2020zeroshot,soh2020zeroshot}. We plan to investigate the effectiveness of network pretraining on the SR accuracy and training time. Second, because recent smartphone devices have multiple cameras with different field-of-views, we expect that helpful reference images can be obtained from other available sources. Last, RZSR uses depth information for reference image retrieval. Other types of information may also be useful, such as semantic segmentation and image salience maps.

\ifCLASSOPTIONcaptionsoff
  \newpage
\fi

\end{document}